\documentclass[conference]{IEEEtran}
\IEEEoverridecommandlockouts
\usepackage{cite}
\usepackage{amsmath, amssymb, amsfonts}
\usepackage{booktabs}
\usepackage{graphics, graphicx, epsfig, subcaption, xcolor}
\usepackage{float}

\renewcommand{\t}{^{\mbox{\tiny\sf T}}}

\renewcommand{\d}{\mathrm{d}}
\newcommand{\dx}{\mathrm{d}x}

\def\BibTeX{{\rm B\kern-.05em{\sc i\kern-.025em b}\kern-.08em
    T\kern-.1667em\lower.7ex\hbox{E}\kern-.125emX}}

\begin{document}
\title{Multi-Contact Force Estimation for Continuum Robots via Gaussian-Parameterized Factor Graphs
\thanks{This work was supported by the AFOSR SURI award FA9550-23-1-0723.}
}

\author{\IEEEauthorblockN{Aditya Prakash}
\IEEEauthorblockA{\textit{School of Aerospace Engineering} \\
\textit{Georgia Institute of Technology}\\
Atlanta, GA, USA \\
leibton@gatech.edu}
\and
\IEEEauthorblockN{Patricio A. Vela}
\IEEEauthorblockA{\textit{School of Electrical and Computer Engineering} \\
\textit{Georgia Institute of Technology}\\
Atlanta, GA, USA \\
pvela@gatech.edu}
\and
\IEEEauthorblockN{Panagiotis Tsiotras}
\IEEEauthorblockA{\textit{School of Aerospace Engineering} \\
\textit{Georgia Institute of Technology}\\
Atlanta, GA, USA \\
tsiotras@gatech.edu}}

\maketitle

%%%%%%%%%%%%%%%%%%%%%%%%%%%%%%%%%%%%%%%%%%%%%%%%%%%%%%%%%%%%%%%%%%%%%%%%%%%%%%%%
\begin{abstract}
Continuum robots offer key advantages in navigating unstructured environments, but their safe operation requires accurate estimation of the external contact forces acting anywhere along the robot body.  
Estimating these forces at unknown locations is an ill-conditioned problem, particularly for multiple contacts. 
We propose a unified shape and force estimation framework formulated on a factor graph.
By incorporating a Gaussian mixture force parameterization into a discretized probabilistic Cosserat rod model, we reduce the dimensionality of the unknown external forces and  mitigate the ill-conditioning of node-wise force estimation.
The framework fuses strain, tendon tension, and pose measurements to simultaneously estimate the robot's shape and external forces while accounting for modeling and sensor uncertainties. 
Numerical simulations demonstrate that the proposed method outperforms existing methods in terms of force location and magnitude estimation for both single and multi-contact scenarios. 
We further present a progressive variant that introduces basis functions on demand to estimate contact forces sequentially during a simulated confined-navigation task.
\end{abstract}

\section{Introduction}

Continuum robots \cite{robinson1999continuum} offer an inherent mechanical advantage over rigid robots in unstructured and confined environments due to their flexibility and compliance \cite{russo2023continuum}.
Continuous deformation makes soft robots well-suited for operations in complex and cluttered environments. 
Furthermore, their structural compliance enables continuum robots to interact safely with the environment without damaging the robot or its surroundings. 
Together, these characteristics enable continuum and soft robots to operate effectively in unstructured and confined environments that are completely unknown or highly uncertain \cite{wu2025soft, abah2021multi}, making them suited for several applications, including minimally invasive surgery \cite{burgner2015continuum}, post-disaster search and rescue \cite{yamauchi2022development}, and the inspection of complex machinery \cite{wang2019aero}. 

Safe and effective navigation in unknown or uncertain environments requires accurate force sensing along the continuum robot \cite{hu2026contact}. For example, force estimates can be used to regulate the contact force between the catheter and tissue during cardiac ablation procedures. 
It can also be used during navigation through constrained paths, such as pipes, to enable adaptive stiffness and shape adjustments, ensuring safe and smooth movement. 
In the absence of an explicit environmental map and visual sensors, the robot must localize and estimate contact forces utilizing only on-board measurements. 

Recent research has addressed force estimation for continuum robots, primarily focusing on isolated tip-force estimation \cite{feng2021learning, diezinger20223d, bian2025accurate, sadati2020stiffness}. 
Estimating forces at the tip is  effective for targeted tasks like ablation or tactile exploration, where the robot body remains in free space while the tip interacts with the environment. 
However, tip-contact assumptions break down during operation in unstructured and unknown environments, where collisions can occur anywhere along the robot's length \cite{hu2026contact}. 

To address collisions along continuum robots, the authors in \cite{bajo2011kinematics} proposed a kinematics-based method for contact detection and localization using screw motion deviation. 
Their method primarily addressed single-contact estimation and was extended to multiple-contact estimation in \cite{job2023multiple} using a contact particle filter and a kinetostatic model. 
However, reliable estimation was confined to the distal segments, and the particle filter scaled poorly as the number of contacts grows. 
Alternatively, in \cite{aloi2022estimating} the authors used a statics model to jointly localize and estimate contact forces. 
They showed that the force estimation problem is inherently ill-posed, as multiple force distributions can yield the same physical deformation, and certain external forces have reduced observability due to the lack of resulting deformation. 
To resolve this, the authors of \cite{aloi2022estimating} used Gaussian functions to parameterize the external load and showed that they can successfully approximate both point loads and uniformly distributed loads by varying the standard deviation of the parameterization.
However, their approach relies on known pose measurements and deterministic models, making it vulnerable to sensor noise. 

In parallel, there has been a trend toward using proprioceptive measurements for force estimation, with a particular focus on strain and tendon tension. 
In \cite{ha2022contact}, contact is localized along a catheter using the shape measured by an embedded multi-core FBG fiber. 
Similarly, \cite{wang2026strain} combines embedded stretch sensors with actuation variables in a kinetostatic strain model to recover both shape and 3D force. 
In \cite{gao2024body}, tension profiles alongside FBG-instrumented actuation fibers are used to recover the location and magnitude of body contacts. 
Furthermore, in \cite{feliu2025actuation}, force and shape are estimated from tendon length and tendon tension alone. 
Together, these studies demonstrate that utilizing backbone strain and tendon tension provides an experimentally proven, sensor-efficient approach to force estimation.

Building on these multimodal sensing strategies, the authors of \cite{ferguson_unified_2024} introduced a unified shape and force estimation framework that utilizes Gaussian process regression. 
While effective, modeling distributed forces via Brownian motion is poorly suited for the localized, discrete point forces characteristic of environmental obstacle collisions. 
Reference \cite{ferguson2026fast} provided the probabilistic foundation needed for modeling point forces by introducing a discretized stochastic Cosserat model that accounts for uncertainty in sensing, tendons, and mechanics. 
The model in \cite{ferguson2026fast} admits a wrench at every node; however, determining point forces at unknown locations would require estimating forces at every node, rendering the estimation problem severely ill-conditioned.

To overcome the previous limitations, we incorporate a Gaussian mixture parameterization of the distributed external forces in the discrete Cosserat model and formulation of a unified shape-and-force estimation problem. 
We exploit the Lie group algebra  structure of the problem formulation to derive the integration of the discretized model for the parameterized forces. 
We then formulate the state estimation problem using a factor graph architecture \cite{dellaert2017factor}, which allows us to probabilistically fuse the multi-sensory measurements with the robot's static model. 
By varying the standard deviation of the Gaussian functions, the parameterization can model both point and distributed forces. 
By controlling the number of basis functions, we reduce the dimensionality of the unknown external forces. 

The rest of the paper is organized as follows: Section~II discusses the Cosserat rod model and its discretized version with force parameterization. Section~III formulates the factor graph for shape and force estimation. 
Section~IV presents numerical simulation results demonstrating the efficiency of the proposed method, followed by Section~V, which demonstrates the framework's practical application for progressively estimating sequential contacts within a confined environment. 
Finally, we conclude the paper in Section~VI.

\section{Cosserat Rod Model}

\begin{figure}[htbp]
    \centering
    \includegraphics[width=1.0\linewidth]{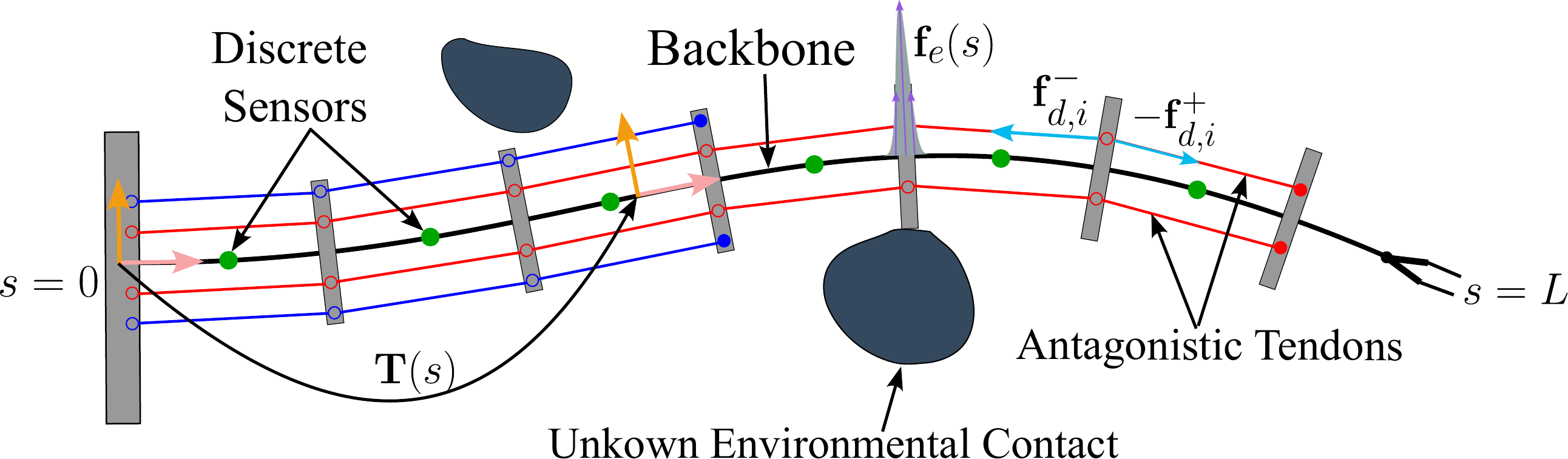}
    \caption{Schematic of a tendon driven continuum robot operating in an unknown environment.}
    \label{fig:force_estimation_schematics}
\end{figure}

The pose of a continuum rod-shaped robot along its arc length, $s \in [0, L]$, is denoted by $\mathbf{T}(\cdot) : s \mapsto \mathbf{T}(s)  \in \mathrm{SE}(3)$. It is typically represented by the matrix
\begin{equation}
    \mathbf{T}(s) = \begin{bmatrix}
        \mathbf{R}(s) & \mathbf{r}(s) \\
        \mathbf{0} & 1
    \end{bmatrix},
\end{equation}
where $\mathbf{r}(s) \in \mathbb{R}^3$ is the position vector of the body frame origin at $s$, and $\mathbf{R}(s) \in \mathrm{SO}(3)$ is the rotation matrix representing the orientation of the body frame at $s$ with respect to the world frame. 
Figure~\ref{fig:force_estimation_schematics} shows the schematic of a continuum robot.
The kinematics equation is given by
\begin{equation}
    \frac{\d }{\d s} \mathbf{T}(s) = \mathbf{T}(s) \xi(s)^\wedge,
    \label{eq:kinematics}
\end{equation}
where $\boldsymbol{\xi}(\cdot): s \mapsto \boldsymbol{\xi}(s) =  \begin{bmatrix}
     \boldsymbol{\Psi}\t & \boldsymbol{\Gamma}\t
\end{bmatrix}\t\in \mathbb{R}^6$ defines the strain of the continuum robot and
\begin{equation}
    \xi(s)^\wedge = \begin{bmatrix}
        \boldsymbol{\Psi}^\times & \boldsymbol{\Gamma} \\
        \mathbf{0} & 0
    \end{bmatrix} \in \mathfrak{se}(3).
\end{equation}
In the above equations, $\boldsymbol{\Psi}(s) \in \mathbb{R}^3$ represents the angular strain, $\boldsymbol{\Gamma}(s) \in \mathbb{R}^3$ represents the linear strain, and $(\cdot)^\times$ denotes the skew-symmetric matrix of the corresponding vector in $\mathbb{R}^3$. 
The constitutive equation is given by
\begin{equation}
    \boldsymbol{\Lambda}(s) = \begin{bmatrix}
        \mathbf{m}(s) \\ \mathbf{n}(s)
    \end{bmatrix} = \mathcal{K}(\boldsymbol{\xi}(s) - \boldsymbol{\xi}^*) + \mathbf{n}_{\bar{\xi}}(s)
    \label{eq:constitutive}
\end{equation}
where $\mathcal{K}$ is the generalized stiffness matrix and $\boldsymbol{\Lambda}(s)$ is the body-frame internal wrench composed of the moment $\mathbf{m}(s)$ and force $\mathbf{n}(s)$. 
In \eqref{eq:constitutive} $\boldsymbol{\xi}^* \in \mathbb{R}^6$ denotes the reference strain of the rod in its natural (unloaded) configuration and $\mathbf{n}_{\bar{\xi}}(s) \sim \mathcal{N}(0, \Sigma_{\bar{\xi}})$ with $\Sigma_{\bar{\xi}} \in \mathbb{R}^{6 \times 6}$ captures the modeling error due to the linear approximation.

The statics of the Cosserat rod in the world frame is governed by \cite{lilge2022continuum}
\begin{equation}
    \frac{\d }{\d s} \boldsymbol{\Lambda}^w(s) = - \mathbf{F}^w_\mathrm{ext}(s) - \mathbf{F}^w_\mathrm{ten}(s),
    \label{eq:statics_world}
\end{equation}
where $\boldsymbol{\Lambda}^w(s) \in \mathbb{R}^6$ is the world-frame internal wrench, related to the body-frame wrench by $\boldsymbol{\Lambda}(s) = \mathrm{Ad}\t(\mathbf{T}(s))\,\boldsymbol{\Lambda}^w(s)$ where the adjoint operator is given by
\begin{equation}
    \mathrm{Ad}(\mathbf{T}) = \begin{bmatrix}
        \mathbf{R} & \mathbf{0} \\
        \mathbf{r}^\times \mathbf{R} & \mathbf{R}
    \end{bmatrix}.
\end{equation}
In \eqref{eq:statics_world}, $\mathbf{F}^w_\mathrm{ext}(s)$ is the distributed load due to external factors, and $\mathbf{F}^w_\mathrm{ten}(s)$ is the distributed load due to tendon actuation, both expressed in the world frame. 
We detail these load models in the subsequent section.

\subsection{Discretization}

We discretize the arc length into $K$ nodes $\{s_k\}_{k=1}^K \subset [0,L]$ with uniform spacing $\Delta s = s_{k+1} - s_k$. 
Assuming that the strain $\boldsymbol{\xi}_k = \boldsymbol{\xi}(s_k)$ is constant over the interval $[s_k,\; s_{k+1}]$, the discrete version of \eqref{eq:kinematics} is
\begin{equation}
    \mathbf{T}_{k+1} = \mathbf{T}_k \exp\!\left( \boldsymbol{\xi}_k^\wedge\, \Delta s\right).
    \label{eq:disc_kinematics}
\end{equation}
We use a midpoint approximation to evaluate the strain ($\boldsymbol{\xi}_k$) from the internal wrench ($\boldsymbol{\Lambda}$). 
From \eqref{eq:constitutive}, we have that
\begin{equation}
    \boldsymbol{\xi}_k = \frac{1}{2} \mathcal{K}^{-1}(\boldsymbol{\Lambda}_k + \boldsymbol{\Lambda}_{k+1}) + \boldsymbol{\xi}^* + \mathbf{n}_{\xi, k},
    \label{eq:disc_midpoint}
\end{equation}
where $\mathbf{n}_{\xi, k} \sim \mathcal{N}(0, \Sigma_\xi)$ with $\Sigma_\xi \in \mathbb{R}^{6 \times 6}$ captures the modeling error due to the linear approximation ($\mathbf{n}_{\bar{\xi}}(s)$) as well as the approximation error due to the discretization.

Because \eqref{eq:statics_world} contains no coupling term in the world frame, it can be integrated exactly over each interval $[s_k, s_{k+1}]$, yielding:
\begin{equation}
\begin{aligned}
    \boldsymbol{\Lambda}^w_{k+1} &= \boldsymbol{\Lambda}^w_k
        - \int_{s_k}^{s_{k+1}}\mathbf{F}^{w}_{\mathrm{ext}}(x)
          \,\mathrm{d}x
        - \int_{s_k}^{s_{k+1}}
          \mathbf{F}^{w}_{\mathrm{ten}}(x) \,\mathrm{d}x.
\end{aligned}
\label{eq:statics_world_discrete_intermediate}
\end{equation}
Rewriting \eqref{eq:statics_world_discrete_intermediate} back into the body frame using $\boldsymbol{\Lambda}_k = \mathrm{Ad}\t (\mathbf{T}_k) \, \boldsymbol{\Lambda}^w_k$, we obtain
\begin{multline}
    \boldsymbol{\Lambda}_{k+1} = \mathrm{Ad}\t(\mathbf{T}_k^{-1}\mathbf{T}_{k+1})\,\boldsymbol{\Lambda}_k
    \\ - \mathrm{Ad}\t(\mathbf{T}_{k+1}) \int_{s_k}^{s_{k+1}}\mathbf{F}^{w}_{\mathrm{ext}}(x)
          \,\mathrm{d}x
       \\ - \mathrm{Ad}\t(\mathbf{T}_{k+1})  \int_{s_k}^{s_{k+1}}
          \mathbf{F}^{w}_{\mathrm{ten}}(x) \,\mathrm{d}x.
          \label{eq:statics_world_discrete_intermediate2}
\end{multline}

Assuming the environment exerts no moment at the point of contact, the distributed external load in the world frame takes the form:
\begin{equation}
    \mathbf{F}^{w}_{\mathrm{ext}}(s) = \begin{bmatrix}
        \mathbf{r}(s) \times \mathbf{f}_e(s) \\ \mathbf{f}_e(s)
    \end{bmatrix},
\end{equation}
where $\mathbf{f}_e(s)$ is the distributed external force acting at $s$ as shown in Fig.~\ref{fig:force_estimation_schematics}.

In \cite{ferguson2026fast}, the authors model the distributed external load as point forces at each node. 
However, directly estimating external force vectors at every discrete node is an ill-conditioned problem due to the under-determined mapping from distributed forces to backbone shape \cite{aloi2022estimating, ferguson2026fast}. 
Therefore, following \cite{aloi2022estimating}, we parameterize the distributed external force in the world frame using a fixed number $M$ of normalized Gaussian basis functions:
\begin{equation}
\begin{aligned}
    \mathbf{f}_e(s) &= \sum_{i = 1}^M \begin{bmatrix} \alpha_{x, i} \\
        \alpha_{y, i} \\ \alpha_{z, i}
    \end{bmatrix}\frac{1}{\sqrt{2\pi}\sigma_i}\;\exp\left(-\frac{(s - \mu_i)^2}{2\sigma_i^2}\right) \\
\end{aligned}
    \label{eq:force_param}
\end{equation}
This force parameterization reduces the dimensionality of the force parameters to be estimated. 
To ensure the basis centers $\mu_i$ remain strictly bounded within $[0, L]$, we apply the parameterization:
\begin{equation}
    \mu_i = h(\beta_i) = \frac{L}{2} \bigl(\tanh(\beta_i) + 1\bigr).
    \label{eq:mu_param}
\end{equation}
Furthermore, let $\boldsymbol{\alpha}_i \triangleq [\alpha_{x,i}, \alpha_{y,i}, \alpha_{z,i}]\t$. 
Then, the force parameter vector to be estimated becomes $\boldsymbol{\gamma} = \bigl[\boldsymbol{\alpha}_1\t, \beta_1, \dots, \boldsymbol{\alpha}_M\t, \beta_M\bigr]\t \in \mathbb{R}^{4M}$ . 
Defining the cumulative external force as
\begin{equation}
\begin{aligned}
    \mathbf{v}_{e,k} &= \int_0^{s_k} \mathbf{f}_e(x)\,\d x \\
    &= \frac{1}{2}\sum_{i=1}^M \begin{bmatrix} \alpha_{x,i} \\ \alpha_{y,i} \\ \alpha_{z,i} \end{bmatrix}\left[\mathrm{erf}\!\left(\frac{s_k - \mu_i}{\sqrt{2}\sigma_i}\right)  - \mathrm{erf}\!\left(\frac{- \mu_i}{\sqrt{2}\,\sigma_i}\right)\right],
\end{aligned}
    \label{eq:v_def}
\end{equation}
it follows that the integral of the distributed external force over the interval $[s_k, s_{k+1}]$ is given by
\begin{equation}
    \int_{s_k}^{s_{k+1}} \mathbf{f}_e(x)\,\dx = \mathbf{v}_{e,k+1} - \mathbf{v}_{e,k}.
\end{equation}
The discrete external wrench contribution in \eqref{eq:statics_world_discrete_intermediate2} can then be evaluated as
\begin{equation}
    \begin{aligned}
        \mathcal{I}_1 &= \mathrm{Ad}\t(\mathbf{T}_{k+1}) \int_{s_k}^{s_{k+1}}\mathbf{F}^{w}_{\mathrm{ext}}(x)
          \,\mathrm{d}x \\
          &= \mathrm{Ad}\t(\mathbf{T}_{k+1}) \begin{bmatrix}
              \int_{s_k}^{s_{k+1}} \mathbf{r}(x) \times \mathbf{f}_e(x) \d x \\
              \int_{s_k}^{s_{k+1}} \mathbf{f}_e(x) \d x
          \end{bmatrix} \\
          &= \mathrm{Ad}\t(\mathbf{T}_{k+1}) \begin{bmatrix}
              \mathbf{r}_{k+1} \times (\mathbf{v}_{e,k+1} - \mathbf{v}_{e,k}) + \mathbf{n}_m\\
              \mathbf{v}_{e,k+1} - \mathbf{v}_{e,k}
          \end{bmatrix},
    \end{aligned}
\end{equation}
where $\mathbf{n}_m = \int_{s_k}^{s_{k+1}} (\mathbf{r}(x) - \mathbf{r}_{k+1}) \times \mathbf{f}_e(x) \d x= \mathcal{O}(\Delta s)$ is the error due to the moment approximation. Expanding the adjoint transformation yields the simplified body-frame formulation:
\begin{equation}
\begin{aligned}
    \mathcal{I}_1 &= \begin{bmatrix}
        \mathbf{R}\t_{k+1} & -\mathbf{R}\t_{k+1} \mathbf{r}_{k+1}^\times \\
         \mathbf{0} & \mathbf{R}\t_{k+1}
    \end{bmatrix} \begin{bmatrix}
              \mathbf{r}_{k+1} \times (\mathbf{v}_{e,k+1} - \mathbf{v}_{e,k}) + \mathbf{n}_m\\
              \mathbf{v}_{e,k+1} - \mathbf{v}_{e,k}
          \end{bmatrix} \\
          &= \mathrm{Rot}\t(\mathbf{T}_{k+1}) \begin{bmatrix}
              \mathbf{n}_m \\
              \mathbf{v}_{e,k+1} - \mathbf{v}_{e,k}
          \end{bmatrix},
\end{aligned}
\label{eq:I1}
\end{equation}
where $\mathrm{Rot}(\mathbf{T}_k) = \mathrm{diag}(\mathbf{R}_k, \mathbf{R}_k)$. 

Similarly, the tendons interact with the backbone only at the routing discs as shown in Fig.~\ref{fig:force_estimation_schematics}. 
Denoting $\mathcal{D} \subseteq \{1, \dots, K\}$ as the set of arc-length indices coincident with a disc, the distributed actuation load is concentrated at these locations
\begin{equation}
    \mathbf{F}^w_\mathrm{ten}(s) = \sum_{d \in \mathcal{D}} \delta(s - s_d)\, \mathbf{F}^w_{\mathrm{ten}, d},
\end{equation}
where $\mathbf{F}^w_{\mathrm{ten}, d}$ is the discrete wrench due to tendon actuation at disc $d$.
Using the above tendon actuation model, we get the discrete tendon contribution in \eqref{eq:statics_world_discrete_intermediate2} to be 
\begin{equation}
    \begin{aligned}
        \mathcal{I}_2 &= \mathrm{Ad}\t(\mathbf{T}_{k+1})  \int_{s_k}^{s_{k+1}}
          \mathbf{F}^{w}_{\mathrm{ten}}(x) \,\mathrm{d}x \\
          &= \mathrm{Ad}\t(\mathbf{T}_{k+1})  \int_{s_k}^{s_{k+1}}
           \sum_{d \in \mathcal{D}} \delta(x - s_d)\, \mathbf{F}^w_{\mathrm{ten}, d} \,\mathrm{d}x \\
           &= \mathbf{1}[k+1 \in \mathcal{D}]\, \mathrm{Ad}\t(\mathbf{T}_{k+1}) \mathbf{F}^w_{\mathrm{ten}, k+1} \\
           &= \mathbf{1}[k+1 \in \mathcal{D}]\, \mathrm{Rot}\t(\mathbf{T}_{k+1}) \mathbf{D}_{k+1}
    \end{aligned}
    \label{eq:I2}
\end{equation}
where $\mathbf{1}[\cdot]$ is the indicator function, $\mathrm{Ad}\t(\mathbf{T}_{k+1}) \mathbf{F}^w_{\mathrm{ten}, k+1} =  \mathrm{Rot}\t(\mathbf{T}_{k+1}) \mathbf{D}_{k+1}$ from \cite{ferguson2026fast},
and
\begin{equation}
    \mathbf{D}_{d} = \sum_{i = 1}^N \left(\mathbf{f}_{d, i}^{-} + \mathbf{f}_{d, i}^{+}\right) + \mathbf{n}_{D, d}, \quad d \in \mathcal{D}.
    \label{eq:tendon_actuation}
\end{equation}
Here, $\mathbf{n}_{D, d}\sim\mathcal{N}(\mathbf{0}, \Sigma_D)$ models small deviations such as tendon-disc friction. 
We refer to \cite{ferguson2026fast} for the detailed modeling of the single-tendon wrenches $\mathbf{f}_{d,i}^{\pm}$ (shown in Fig.~\ref{fig:force_estimation_schematics}), which depend on the tendon routing geometry and the tendon tension vector.
Substituting \eqref{eq:I1} and \eqref{eq:I2} in \eqref{eq:statics_world_discrete_intermediate2} yields
\begin{multline}
    \boldsymbol{\Lambda}_{k+1} = \mathrm{Ad}\t(\mathbf{T}_k^{-1}\mathbf{T}_{k+1})\,\boldsymbol{\Lambda}_k
    \\ -  \mathrm{Rot}\t(\mathbf{T}_{k+1}) \begin{bmatrix}
            \mathbf{0} \\
              \mathbf{v}_{e,k+1} - \mathbf{v}_{e,k}
          \end{bmatrix}
       \\ -  \mathbf{1}[k+1 \in \mathcal{D}]\, \mathrm{Rot}\t(\mathbf{T}_{k+1}) \mathbf{D}_{k+1} + \mathbf{n}_{\Lambda, k},
       \label{eq:wrench_discrete}
\end{multline}
where the process noise $\mathbf{n}_{\Lambda, k} = [\mathbf{n}_m\t, \mathbf{n}_n\t]\t \in \mathbb{R}^6$ absorbs the interval approximation error. 
The moment component $\mathbf{n}_m$ explicitly captures the $\mathcal{O}(\Delta s)$ lumping error derived in \eqref{eq:I1}, while the force component $\mathbf{n}_n$ is drawn from a zero-mean Gaussian with a very small variance to softly enforce the static balance constraint during estimation.

\section{Problem Formulation}

We consider the problem of unified shape and external force estimation for a continuum robot using the discretized Cosserat rod model described above, given measurements at discrete arc-length locations and known tendon tensions.

\begin{figure*}
    \centering
    \includegraphics[width=1.0\linewidth]{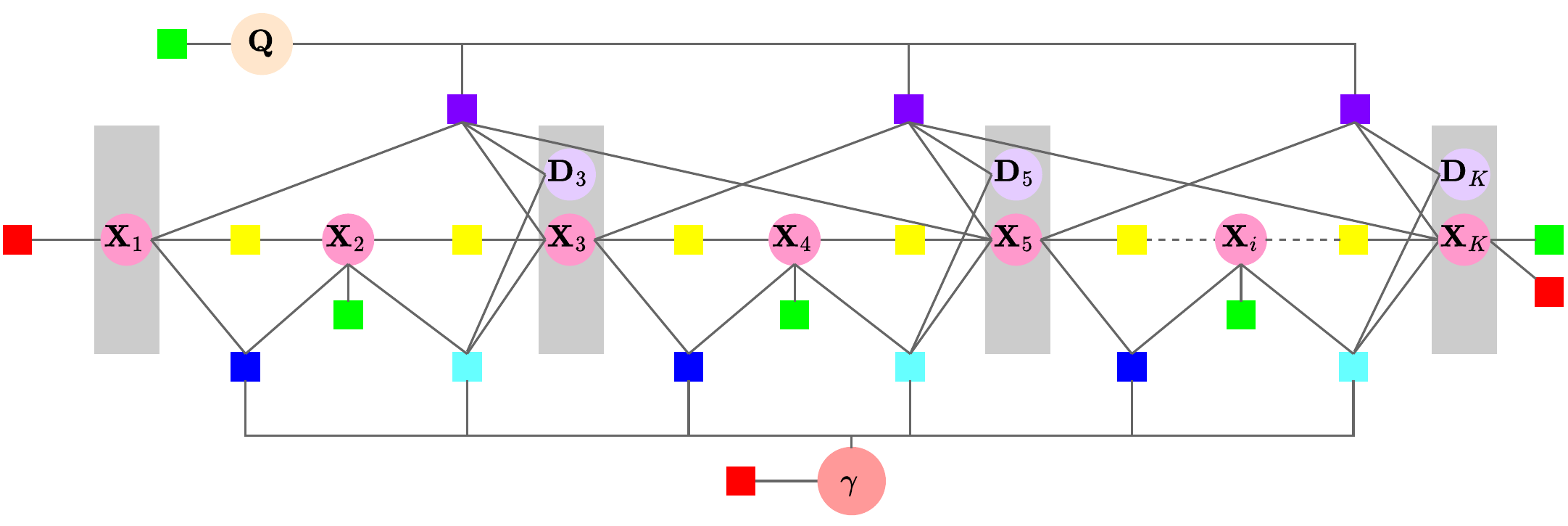}
    \caption{Diagram of the proposed unified shape and force estimation factor graph. The circles represent the unknowns whereas the squares represent the factors. 
    The green squares represent the measurement factors, the red squares represent the prior factors, the yellow squares represent the combined kinematics/constitutive factors, the violet squares represent the tendon actuation factors, and the blue and cyan squares represent the wrench balance factors (where cyan specifically denotes a node that coincides with a routing disc).}
    \label{fig:factor_graph}
\end{figure*}

In our discretized setting, the unknowns are the poses $\mathbf{T}_k \in \mathrm{SE}(3)$,  and the internal wrenches $\boldsymbol{\Lambda}_k \in \mathbb{R}^6$ at each node $k \in \{1,\ldots,K\}$. 
Additionally, we estimate the disc-specific concentrated tendon wrench states $\mathbf{D}_d \in \mathbb{R}^6$ at each routing disc $d \in \mathcal{D}$, the tendon tension vector $\mathbf{Q} \in \mathbb{R}^N$, and the external force parameter vector $\boldsymbol{\gamma} \in \mathbb{R}^{4M}$, which encodes the amplitude and location of each of the $M$ Gaussian basis functions.
Collecting node-wise states into $\mathbf{X}_k = \{\mathbf{T}_k, \boldsymbol{\Lambda}_k\}$, the full unknown state vector is defined as
\begin{equation}  \label{param:eqn}
    \boldsymbol{\Theta} = 
    \Bigl(\{\mathbf{X}_k\}_{k = 1}^K,\{\mathbf{D}_{d}\}_{d\in\mathcal{D}}, \mathbf{Q}, \boldsymbol{\gamma}\Bigr).
\end{equation}

\subsection{Factor Graph Construction}

We adopt the factor graph framework \cite{dellaert2017factor} to formulate a Maximum a Posteriori (MAP) estimation problem in terms of the parameter vector in \eqref{param:eqn}.
A factor graph is a bipartite graphical model that factors the joint posterior of the entire system into a product of local factors, each constraining a subset of the variables. 
It consists of two types of nodes: \textit{variable nodes} representing the unknown states to be estimated (in our case, the combined state vector $\boldsymbol{\Theta}$), and \textit{factor nodes} representing the physical or measurement constraints applied to those variables. 
Under Gaussian noise assumptions, this formulation reduces the MAP estimation to a nonlinear least-squares optimization~\cite{barfoot2024state}.

To comprehensively define our estimation problem, we group the factor nodes into three distinct categories: mechanics factors, measurement factors, and prior boundary factors. 
Fig.~\ref{fig:factor_graph} illustrates the architecture of the proposed factor graph, with variable nodes represented by circles and factor nodes by squares.

\subsubsection{Mechanics Factors}

The first category consists of three sets of factors that enforce the discretized Cosserat rod model derived in Section II.

\textit{Kinematics factor.} By combining the discrete kinematic equation~\eqref{eq:disc_kinematics} and the constitutive midpoint approximation~\eqref{eq:disc_midpoint}, we enforce strain-pose consistency. The residual vector at interval $k$ is defined as
\begin{equation}
    \mathbf{e}_{\xi,k} = \frac{1}{2}\mathcal{K}^{-1}(\boldsymbol{\Lambda}_k + \boldsymbol{\Lambda}_{k+1}) + \boldsymbol{\xi}^* - \frac{1}{\Delta s} \log\left(\mathbf{T}_k^{-1}\mathbf{T}_{k+1}\right)^\vee.
\end{equation}
In Fig.~\ref{fig:factor_graph}, this factor is represented by the yellow squares. It yields the cost term
\begin{equation}
    J_{\xi,k} = \frac{1}{2}\,\mathbf{e}_{\xi,k}\t\,
                \Sigma_\xi^{-1}\,\mathbf{e}_{\xi,k}.
    \label{eq:kin_factor}
\end{equation}

\textit{Wrench balance factor.} From the discrete body-frame static balance equation~\eqref{eq:wrench_discrete}, the residual at interval $k$ is
\begin{multline}
    \mathbf{e}_{\Lambda,k} = \boldsymbol{\Lambda}_{k+1} 
    - \mathrm{Ad}\t \bigl(\mathbf{T}_k^{-1}\mathbf{T}_{k+1}\bigr)\boldsymbol{\Lambda}_k \\
    + \mathrm{Rot}\t (\mathbf{T}_{k+1})
      \begin{bmatrix}\mathbf{0} \\ \mathbf{v}_{e,k+1} - \mathbf{v}_{e,k}\end{bmatrix} \\
    + \mathbf{1}_{[k+1\in\mathcal{D}]}\,\mathrm{Rot}\t(\mathbf{T}_{k+1}) \mathbf{D}_{k+1},
    \label{eq:wrench_residual}
\end{multline}
where $\mathbf{v}_{e,k}$ is defined in~\eqref{eq:v_def} and depends on the force parameters $\boldsymbol{\gamma}$. 
The concentrated tendon actuation term $\mathbf{D}_{k+1}$ is added via the indicator function only when the node coincides with a routing disc. 
In Fig.~\ref{fig:factor_graph}, the blue squares represent this factor at standard nodes, and the cyan squares at disc nodes. 
The associated cost is
\begin{equation}
    J_{\Lambda,k} = \frac{1}{2}\,\mathbf{e}_{\Lambda,k}\t\,
                    \Sigma_\Lambda^{-1}\,\mathbf{e}_{\Lambda,k}.
    \label{eq:wrench_factor}
\end{equation}

\textit{Tendon actuation factor.} We enforce the analytical tendon actuation relationship originating from the tendon routing geometry. 
From~\eqref{eq:tendon_actuation}, the residual at disc $d$ is
\begin{equation}
    \mathbf{e}_{D, d} = \mathbf{D}_{d} - \sum_{i=1}^{N}\left(\mathbf{f}_{d,i}^{-} + \mathbf{f}_{d,i}^{+}\right),
\end{equation}
with the associated cost function being
\begin{equation}
    J_{D,d} = \frac{1}{2}\,\mathbf{e}_{D,d}\t\,
                    \Sigma_D^{-1}\,\mathbf{e}_{D,d}.
\end{equation}
This factor is represented by the violet squares in Fig.~\ref{fig:factor_graph}. 

\subsubsection{Measurement Factors}

Measurement factors ground the robot's internal states in observable, real-world data, rendering the otherwise underdetermined mechanics into a well-posed estimation problem.
The various measurement modalities are given below and are represented by the green squares in Fig.~\ref{fig:factor_graph}. 

\textit{Tendon tension measurement.} To distinguish external wrenches from internal actuation loads, the tendon tensions must be measured. Given the observed tension vector $\bar{\mathbf{Q}}$, the residual is
\begin{equation}
    \mathbf{e}_{Q} = \mathbf{Q} - \bar{\mathbf{Q}},
\end{equation}
yielding the cost term
\begin{equation}
    J_Q = \frac{1}{2}\,\mathbf{e}_{Q}\t\,\Sigma_Q^{-1}\,\mathbf{e}_{Q},
    \label{eq:tension_factor}
\end{equation}
where $\Sigma_Q$ is the measurement covariance.

\textit{Pose measurements.} Let $\mathcal{S}_T \subseteq \{1, \ldots, K\}$ denote the set of node indices where the local pose (or part thereof) is observed by sensors (e.g., electromagnetic trackers). 
At each node $k \in \mathcal{S}_T$, the measured pose is $\bar{\mathbf{T}}_k \in \mathrm{SE}(3)$, and the corresponding residual is formulated as
\begin{equation}
    \mathbf{e}_{\mathrm{T},k} = \log\!\left(\bar{\mathbf{T}}_k^{-1}\mathbf{T}_k\right)^\vee,
\end{equation}
yielding the cost term
\begin{equation}
    J_{\mathrm{T},k} = \frac{1}{2} \mathbf{e}_{\mathrm{T},k}\t \Sigma_{T_m}^{-1} \mathbf{e}_{\mathrm{T},k}.
    \label{eq:pose_factor}
\end{equation}

\textit{Strain measurements.} Similarly, let $\mathcal{S}_\xi \subseteq \{1, \ldots, K\}$ denote the set of node indices where local strain (or part thereof) is observed (e.g., via FBG sensors). 
At each node $k \in \mathcal{S}_\xi$, given the measured strain $\bar{\xi}_k$, the residual is
\begin{equation}
    \mathbf{e}_{\xi_m,k} = \mathcal{K}^{-1} \boldsymbol{\Lambda}_k + \boldsymbol{\xi}^*- \bar{\xi}_k,
\end{equation}
yielding the cost term
\begin{equation}
    J_{\xi_m,k} = \frac{1}{2} \mathbf{e}_{\xi_m,k}\t \Sigma_{\xi_m}^{-1} \mathbf{e}_{\xi_m,k}.
    \label{eq:strain_factor}
\end{equation}

\subsubsection{Prior and Boundary Condition Factors}
Priors and boundary conditions encode the system's structural constraints and regularize the otherwise ill-posed inverse problem. 
In Fig.~\ref{fig:factor_graph}, these are represented by the red squares. 

\textit{Base boundary condition.} The robot's base pose $\bar{\mathbf{T}}_0$ is structurally constrained. The residual is
\begin{equation}
    \mathbf{e}_{\mathrm{base}} = \log\!\left(\bar{\mathbf{T}}_0^{-1}\mathbf{T}_1\right)^\vee,
\end{equation}
yielding the cost term
\begin{equation}
    J_{\mathrm{base}} = \frac{1}{2} \mathbf{e}_{\mathrm{base}}\t \Sigma_0^{-1} \mathbf{e}_{\mathrm{base}},
    \label{eq:base_bc}
\end{equation}
with $\Sigma_0 = \mathrm{diag}(\sigma_p^2 I_3,\, \sigma_\theta^2 I_3)$ set to a small variance to pin the base.

\textit{Tip boundary condition.} The internal wrench at the tip must balance any load applied there. Given a known tip wrench $\mathbf{F}_{\mathrm{tip}}$, the residual is
\begin{equation}
    \mathbf{e}_{\mathrm{tip}} = \boldsymbol{\Lambda}_K - \mathrm{Rot}\t(\mathbf{T}_K)\,\mathbf{F}_{\mathrm{tip}},
\end{equation}
yielding the cost term
\begin{equation}
    J_{\mathrm{tip}} = \frac{1}{2} \mathbf{e}_{\mathrm{tip}}\t \Sigma_K^{-1} \mathbf{e}_{\mathrm{tip}}.
    \label{eq:tip_bc}
\end{equation}
For free-space operation, $\mathbf{F}_{\mathrm{tip}}$ is set to $\mathbf{0}$. 
If tip contact is present, it can instead be supplied by a tip-mounted load cell or promoted to an unknown variable to be estimated jointly.

\textit{Force parameter prior.} We place a zero-mean prior 
$(\boldsymbol{\alpha}_1 = \boldsymbol{\alpha}_2 = \ldots = \boldsymbol{\alpha}_M = \mathbf{0})$
on the Gaussian force amplitudes to regularize the ill-posed mapping between continuous shape and distributed load:
\begin{equation}
    \mathbf{e}_\gamma = \boldsymbol{\alpha}, \quad \text{where} \quad \boldsymbol{\alpha} = [\boldsymbol{\alpha}_{1}\t,\ldots,\boldsymbol{\alpha}_{M}\t]\t,
\end{equation}
yielding the cost term
\begin{equation}
    J_\gamma = \frac{1}{2}\,\mathbf{e}_\gamma\t\,\Sigma_\gamma^{-1}\,\mathbf{e}_\gamma,
    \label{eq:force_prior}
\end{equation}
where $\Sigma_\gamma$ encodes the expected maximum force scale.

\subsection{MAP Objective}

Combining all previous components, the MAP estimate $\hat{\boldsymbol{\Theta}}$ is the solution to the nonlinear least-squares optimization
\begin{equation}
    \hat{\boldsymbol{\Theta}} = \underset{\boldsymbol{\Theta}}{\arg\min}\; J(\boldsymbol{\Theta})
    \label{eq:optim_eq}
\end{equation}
where
\begin{equation}
\begin{aligned}
    J(\boldsymbol{\Theta}) &= 
    \underbrace{J_{\mathrm{base}} + J_{\mathrm{tip}} 
    + J_\gamma}_{J_{\mathrm{prior}}}\\
    &\quad +\underbrace{\sum_{k=1}^{K-1}\!
    \left(J_{\xi,k} + J_{\Lambda,k}\right) + \sum_{d \in \mathcal{D}} J_{D,d}}_{J_{\mathrm{mech}}}\\
    &\quad +\underbrace{\sum_{k\in\mathcal{S}_T}\!J_{\mathbf{T},k}
    + \sum_{k\in\mathcal{S}_\xi}\!J_{\xi_m,k} + J_Q}_{J_{\mathrm{meas}}}.
\end{aligned}
\label{eq:map_cost}
\end{equation}
Here $J(\boldsymbol{\Theta})$ equals the negative log of the joint posterior up to an additive constant; its global minimizer therefore coincides with the MAP estimate of the factor graph. 
We implement the graph in the GTSAM library \cite{dellaert2012factor}, adopting the linearization scheme of \cite{ferguson2026fast} together with the analytical Jacobian with respect to the force parameter vector $\boldsymbol{\gamma}$ derived in Appendix A, and optimize using Powell's Dog-Leg algorithm \cite{powell1970hybrid}. 
We extract the Laplace approximation at the MAP solution to evaluate the posterior uncertainty.

Solving the optimization in problem \eqref{eq:optim_eq} requires an initial state guess. For isolated, single-frame estimation, we initialize the graph variables using the theoretical unloaded configuration of the robot. 
However, during continuous operation across sequential time steps, we instead initialize each step with the optimized posterior $\hat{\boldsymbol{\Theta}}$ from the previous step.

\section{Numerical Simulation Results}

We evaluated the proposed estimator across several force estimation scenarios. 
Ground truth test scenarios were generated using SoroSim \cite{mathew2025reduced}, applying point loads such that their axial force component is zero in the body frame and no tendon actuation. 
The simulated robotic manipulator maintains properties consistent with those established in \cite{aloi2022estimating} to enable direct comparison. 
Furthermore, because the continuum robot is slender, the linear strain is modeled as a constant $[1,\, 0,\, 0]\t$, meaning the backbone is effectively inextensible and unshearable, allowing only the angular strain to vary.  These properties are listed in Table~\ref{tab:properties_manip}. 

\begin{table}[H]
\renewcommand{\arraystretch}{1.5}
    \centering
    \caption{Properties of the continuum manipulator.}
    \label{tab:properties_manip}
    \begin{tabular}{l|l}
    \hline \hline
    Properties & Value\\
    \hline
        Length of the Manipulator & $40 \;\mathrm{cm}$ \\
        Young's Modulus & $207\; \mathrm{GPa}$ \\
        Diameter & $1.4\; \mathrm{mm}$  \\ \hline
    \end{tabular}
\end{table}

For the proposed estimator, the standard deviation $\sigma$ for the Gaussian basis functions is fixed at $12 \; \mathrm{mm}$ to accurately model localized point forces. 
Decreasing $\sigma$ further makes the basis functions unobservable depending on the step size, $\Delta s$, whereas increasing $\sigma$ fails to effectively model highly localized forces. 
Sensor measurements consist of the 3D position of the tip and angular strain measurements taken at 10 evenly spaced locations along the backbone. 
For the noisy simulations, $1\, \mathrm{mm}$ of Gaussian noise is added to the tip position, and $0.1 \;\mathrm{rad\,m}^{-1}$ is added to the angular strain measurements. We benchmark our proposed method against two recent frameworks:

\begin{itemize}
    \item \textbf{Baseline 1} \cite{ferguson2026fast}: A discrete node-wise estimator that estimates external wrench at every node. We run the implementation in \cite{ferguson2026fast} on SoroSim-generated datasets to directly compare performance. 
    To localize the forces using this method, we identify the peaks of the node-wise force profile (one peak for a single force, two for two forces, etc).
    
    \item \textbf{Baseline 2} \cite{aloi2022estimating}: Gaussian force parameterization is used with a deterministic statics model, relying solely on discrete position measurements. 
    Performance metrics for this baseline were taken directly from the original publication. 
    Because the authors evaluated only single-contact scenarios, this baseline is used only for single-force analysis.
\end{itemize}

\subsection{Single Force Estimation}

For the single force estimation analysis, we evaluated 2,000 randomized samples. 
Ground truth forces were randomly generated between $0.5$ and $1.0\; \mathrm{N}$ with varying direction and randomly located between $10 \; \mathrm{cm}$ and $38 \; \mathrm{cm}$ along the backbone, maintaining consistency with \cite{aloi2022estimating}. Table~\ref{tab:single_force} summarizes the results.

\begin{table}[H]
\centering
\renewcommand{\arraystretch}{1.2}
\caption{Simulation results for single force estimation.}
\label{tab:single_force}
\begin{tabular}{lrrr}
\toprule
 & Baseline 1 & Baseline 2 & Proposed \\
\midrule
\multicolumn{4}{l}{\textit{No measurement noise}} \\
Location Error      & 9 mm     & 3.5 mm  & 3.47 mm  \\
Force Spread        & 50.1 mm  & --      & 12 mm    \\
Shape Error         & 0.17 mm  & --      & 0.46 mm  \\
Force Mag. Error    & --       & 0.08 N  & 0.06 N \\
Mean Runtime        & 90.8 ms  & --      & 90.3 ms  \\
Mean No. of Iterations & 9 & --& 23  \\
\midrule
\multicolumn{4}{l}{\textit{With measurement noise}} \\
Location Error      & 36.68 mm & 20.1 mm & 8.74 mm  \\
Force Spread        & 122.59 mm& --      & 12 mm    \\
Shape Error         & 1.15 mm  & --      & 1.34 mm  \\
Force Mag. Error    & --       & 0.53 N  & 0.14 N \\
Mean Runtime        & 122.20 ms & --      & 132.30 ms \\
Mean No. of Iterations & 11 & --& 35  \\
\bottomrule
\end{tabular}
\end{table}

In the noise-free case, the proposed method achieves a mean location error of $3.47 \;\mathrm{mm}$, which is comparable to the $3.5 \; \mathrm{mm}$ of Baseline~2, while significantly outperforming Baseline~1, which has error of $9 \; \mathrm{mm}$. 
When sensor noise is introduced, the proposed method's mean location error remains highly robust at $8.74 \; \mathrm{mm}$, whereas Baseline~2 degrades significantly to $20.1 \; \mathrm{mm}$ and Baseline~1 to $36.68 \; \mathrm{mm}$. 
We observe that while the proposed framework yields results similar to Baseline~2 in the absence of noise, it substantially outperforms Baseline~2 under noisy conditions because the proposed factor graph formulation explicitly models measurement uncertainty. 

To understand why Baseline~1 performs significantly worse than the proposed method, we analyzed the force spread, which is the spatial spread of the estimated force profile about its peak
\begin{equation}
    \sqrt{\sum_k w_k (s_k - s_\mathrm{peak})^2},
\end{equation}
where $w_k$ is the normalized force magnitude at node $k$. 
A small value indicates a sharply localized estimate, while a large value indicates a force smeared across many nodes. 
Because per-node force estimation is inherently ill-conditioned, Baseline~1 spreads the load widely ($50.1 \, \mathrm{mm}$ vs. $12\, \mathrm{mm}$ in the noise-free case), misrepresenting localized contacts. This is reflected in the shape error. 
The over-parameterization in Baseline~1 closely matches the backbone by distributing force across many nodes and, under measurement noise, fitting the noise itself, producing large location errors and a spread in force. 
The proposed Gaussian basis cannot absorb error this way. 
Instead, it keeps the force estimate localized and leaves a small residual in the shape fit.

The magnitude estimate for the proposed method is comparable to Baseline~2 without noise, but significantly superior when noise is introduced. 
Based on these simulation results, the proposed method typically requires more iterations to converge, resulting in a slightly higher mean runtime than 
Baseline 1.
However, per-step time is shorter for the proposed method due to a reduction in the number of unknown variables, despite a loss of sparsity. 

\subsection{Two Force Estimation}

For the two-force estimation case, we evaluated 4,000 randomized samples with ground-truth force magnitudes ranging from 0.3 to $0.8\; \mathrm{N}$. Table~\ref{tab:two_force} reports the comparative results against Baseline~1.

\begin{table}[]
\centering
\renewcommand{\arraystretch}{1.2}
\caption{Simulation results for two force estimation.}
\label{tab:two_force}
\begin{tabular}{lrr}
\toprule
 & Baseline 1 & Proposed \\
\midrule
\multicolumn{3}{l}{\textit{No measurement noise}} \\
Min Location Error   & 7.67 mm  & 3.94 mm  \\

Max Location Error   & 48 mm    & 36.95 mm \\
Avg Location Error   & 27.84 mm & 20.44 mm \\
Shape Error          & 0.17 mm  & 0.91 mm  \\
Avg Force Mag. Error & --       & 0.25 N \\
Mean Runtime         & 86.60 ms  & 93.10 ms  \\
Mean No. of Iterations & 9 & 24  \\
\midrule
\multicolumn{3}{l}{\textit{With measurement noise}} \\
Min Location Error   & 19.29 mm & 9.86 mm  \\
Max Location Error   & 82.47 mm & 48.32 mm \\
Avg Location Error   & 50.88 mm & 29.08 mm \\
Shape Error          & 1.20 mm   & 1.43 mm  \\
Avg Force Mag. Error & --       & 0.27 N \\
Mean Runtime         & 105.60 ms & 140.20 ms \\
Mean No. of Iterations & 11 & 38  \\
\bottomrule
\end{tabular}
\end{table}

When analyzing the average minimum location error, both methods exhibit errors comparable to their single-force estimation performance, indicating that isolating at least one force works well under both clean and noisy conditions. 
However, the average maximum location error indicates that Baseline~1 struggles to localize the second force accurately. 
Under noisy conditions, the maximum location error is bounded at $48.32 \; \mathrm{mm}$ for the proposed estimator, whereas it spikes to $82.47 \; \mathrm{mm}$ for Baseline~1. 

\begin{figure}[htbp]
    \centering
    \begin{subfigure}[b]{0.48\textwidth}
        \centering
        \includegraphics[width=\textwidth]{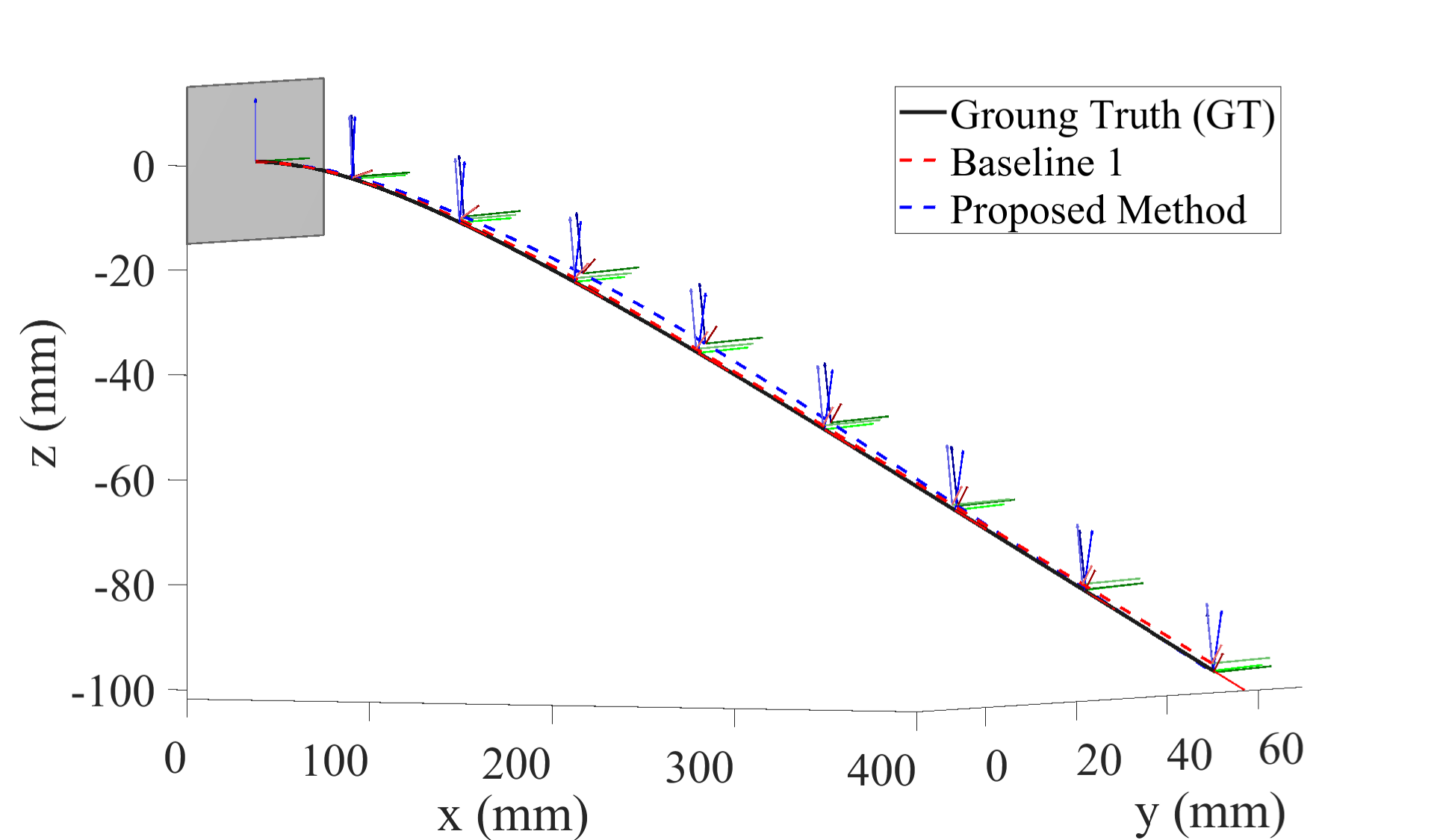}
        \caption{Both methods recover the backbone shape accurately.}
        \label{fig:two_force_profile_shape}
    \end{subfigure}
    \hfill
    \begin{subfigure}[b]{0.48\textwidth}
        \centering
        \includegraphics[width=\textwidth]{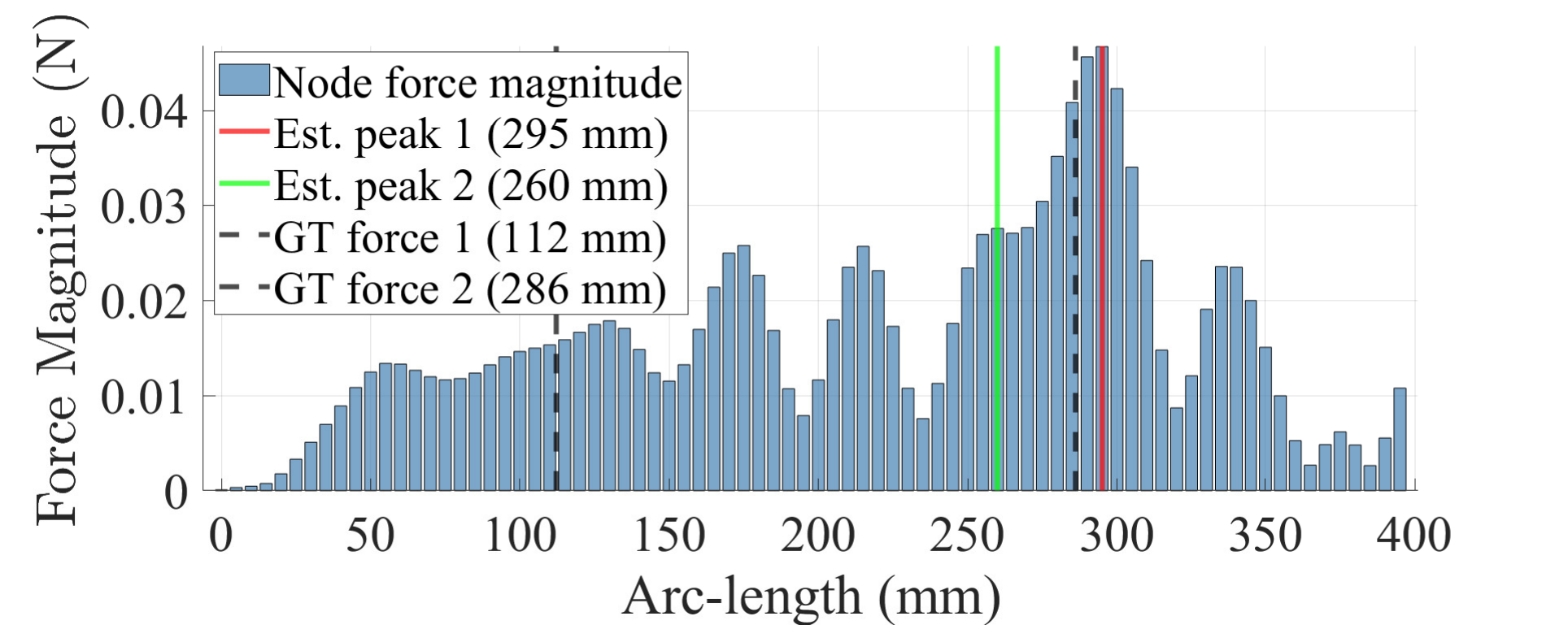}
        \caption{Estimated force profile for Baseline~1.}
        \label{fig:two_force_profile_pfe}
    \end{subfigure}
    \hfill
    \begin{subfigure}[b]{0.48\textwidth}
        \centering
        \includegraphics[width=\textwidth]{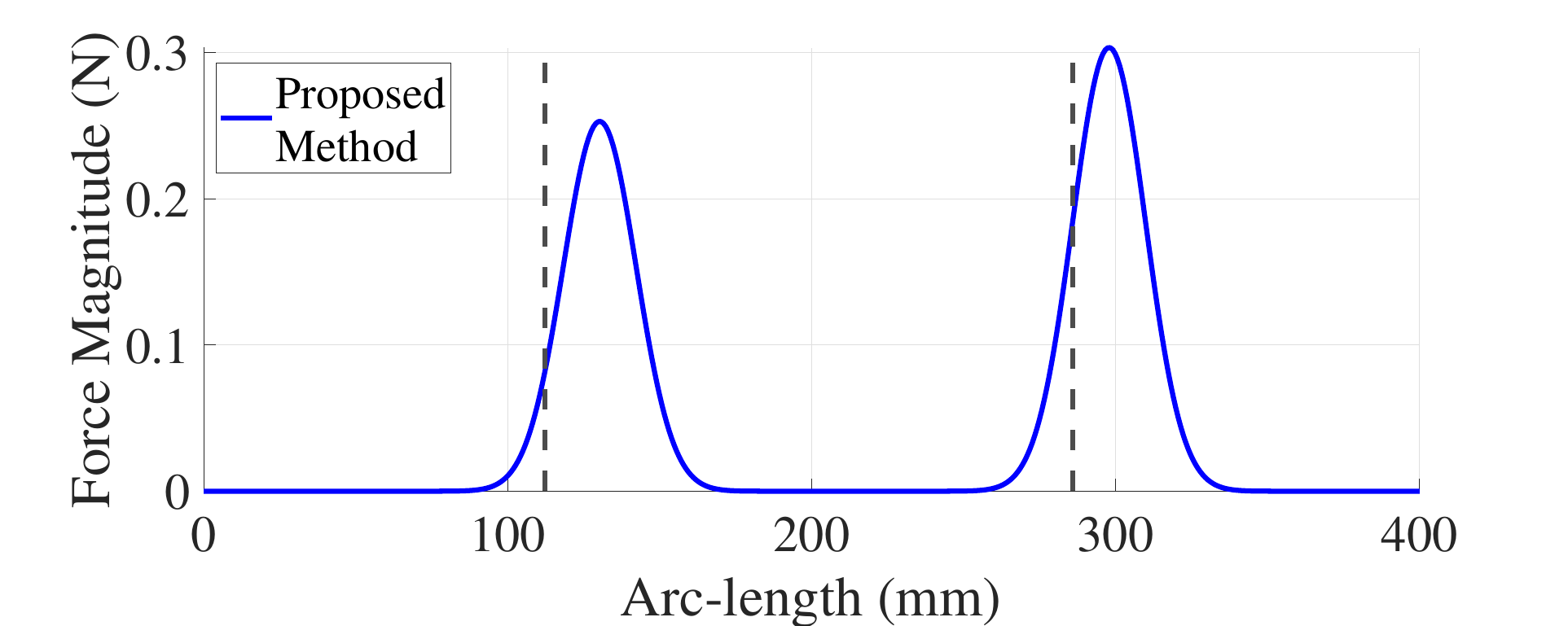}
        \caption{Estimated force profile for the proposed method.}
        \label{fig:two_force_profile_gbfe}
    \end{subfigure}
    \caption{Two force estimation. The black dotted line represents the ground truth force locations.}
    \label{fig:two_force_profile}
\end{figure}

\begin{figure}[htbp]
    \centering
    % First Subfigure
    \begin{subfigure}[b]{0.4\textwidth}
        \centering
        \includegraphics[width=\textwidth]{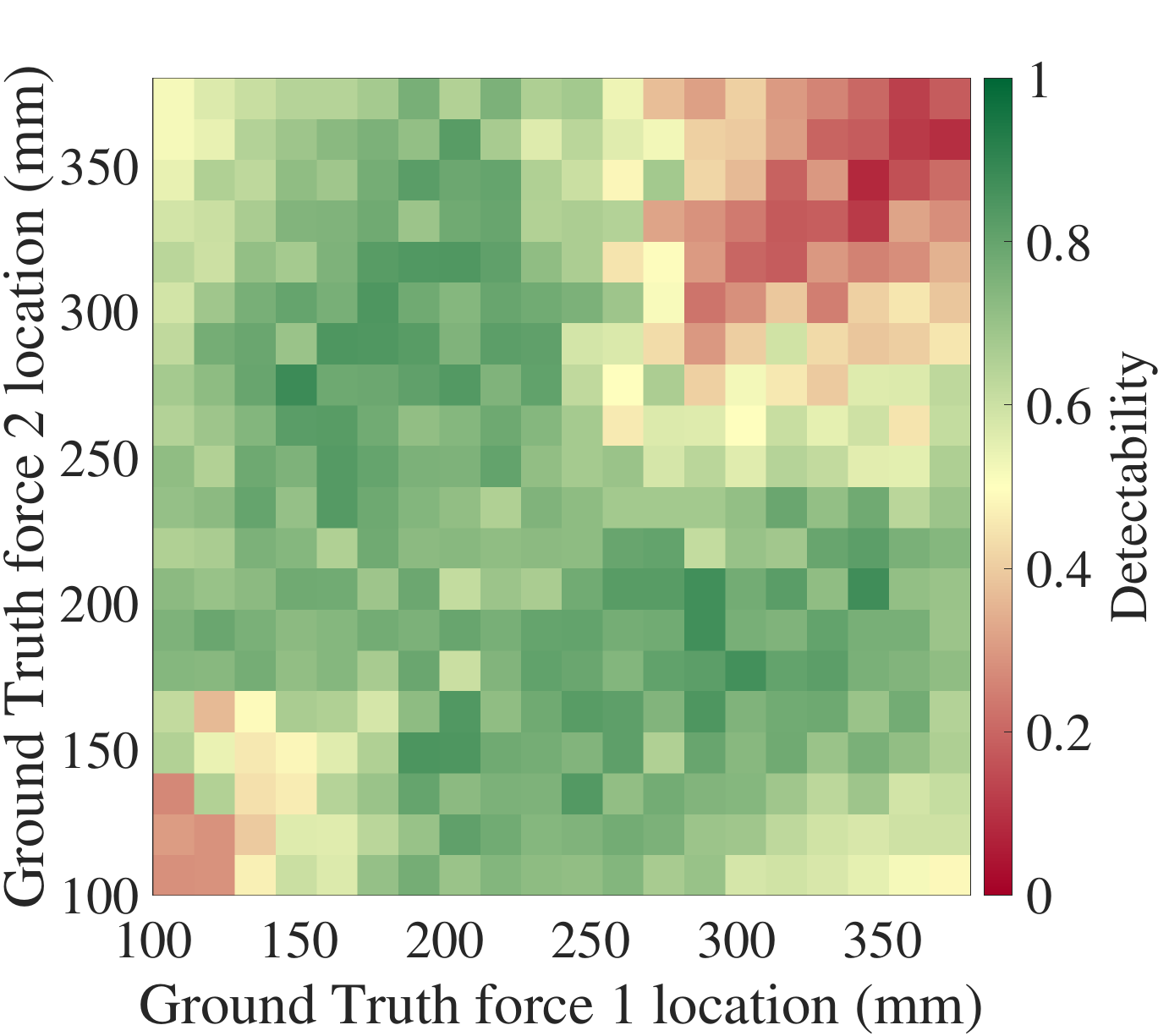}
        \caption{Heatmap showing detectability of forces with ground truth force locations. 0 means only one basis function is detectable, while 1 means both basis functions are detectable.}
        \label{fig:two_force_heatmap_detectability}
    \end{subfigure}
    \hfill % Adds flexible space between the images
    % Second Subfigure
    \begin{subfigure}[b]{0.4\textwidth}
        \centering
        \includegraphics[width=\textwidth]{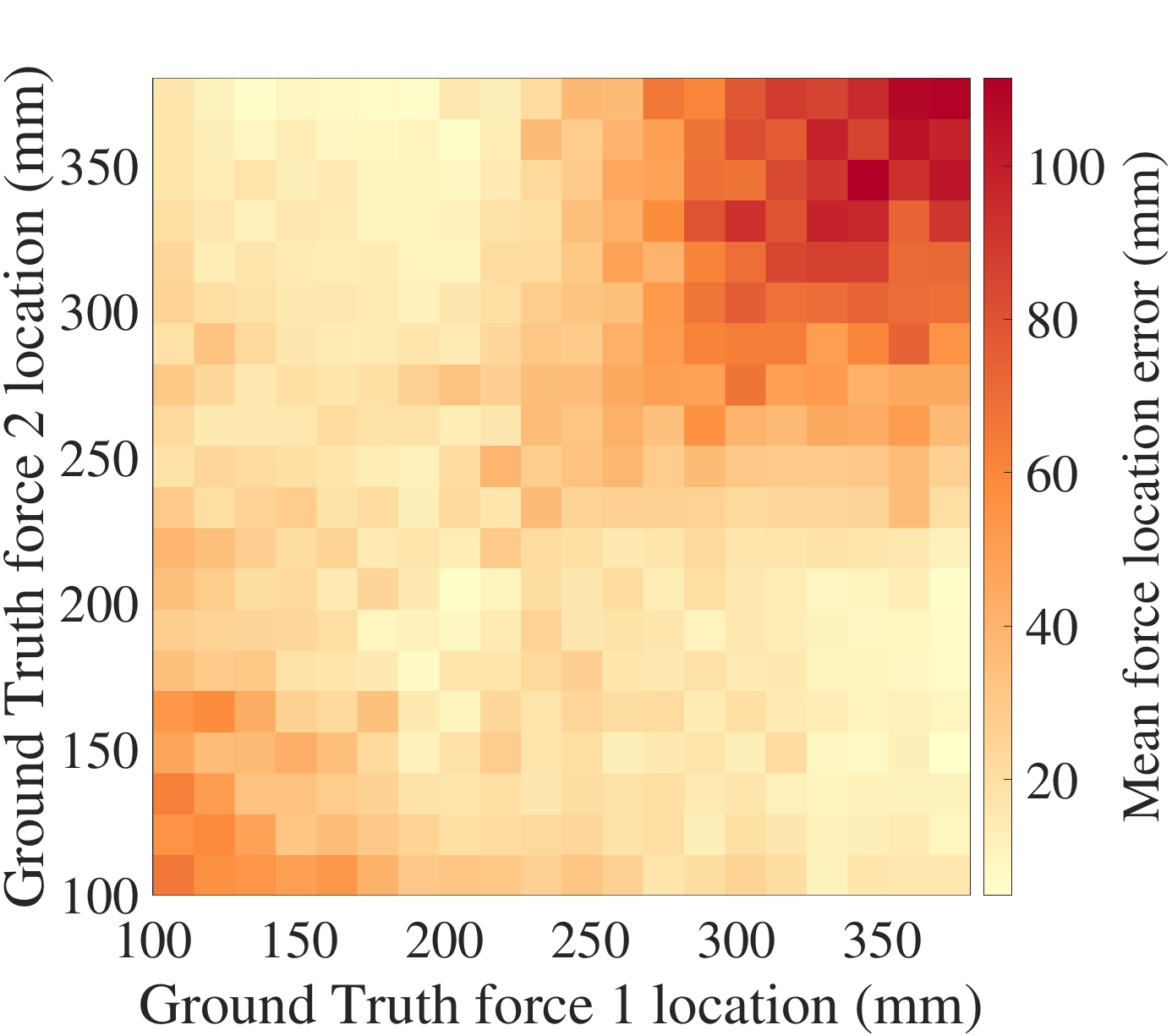}
        \caption{Heatmap showing force location error with ground truth force locations}
        \label{fig:two_force_heatmap_error}
    \end{subfigure}
    
    \caption{Two force estimation analysis for proposed method.}
    \label{fig:two_force_heatmap}
\end{figure}

As in the single-force scenario, Baseline~1 achieves a smaller shape error at the cost of producing a less localized and less accurate force estimate.
Figure~\ref{fig:two_force_profile} makes this contrast visually, illustrating a specific test case where a force of $[0, -0.2358, -0.3003]\t \;\mathrm{N}$ is applied at $112 \;\mathrm{mm}$, and a second force of $[0, 0.2282, -0.2599]\t \;\mathrm{N}$ is applied at $286 \;\mathrm{mm}$. 
While the estimated backbone shapes of both methods track the ground truth closely (Fig.~\ref{fig:two_force_profile_shape}), the recovered force profiles differ sharply. For Baseline~1 (Fig.~\ref{fig:two_force_profile_pfe}), the estimated force spreads across many nodes and introduces several spurious peaks, leading to incorrect force localization.
Conversely, the proposed estimate (Fig.~\ref{fig:two_force_profile_gbfe}) resolves two distinct Gaussian peaks at the correct locations, recovering the two forces directly. 

Figure~\ref{fig:two_force_heatmap} maps the operational boundaries of the proposed method. Detectability (Fig.~\ref{fig:two_force_heatmap_detectability}) is defined as the activation ratio of the basis function magnitudes normalized by the ground truth force magnitude. 
We observe high mean force location errors in specific regions where detectability drops. 
This primarily occurs when both forces are applied in close proximity to one another, particularly near the tip. 
In these highly coupled regions, one basis function absorbs the aggregate effect of both forces, driving the detectability of the second basis function down (Fig.~\ref{fig:two_force_heatmap_detectability}). 
Because the location of this unactivated second basis function is still factored into the aggregate calculation, the mean location error artificially spikes in these configurations.

\section{Progressive Multi-Contact Estimation during Confined Navigation}

\begin{figure}[htbp]
    \centering
    \includegraphics[width=0.8\linewidth]{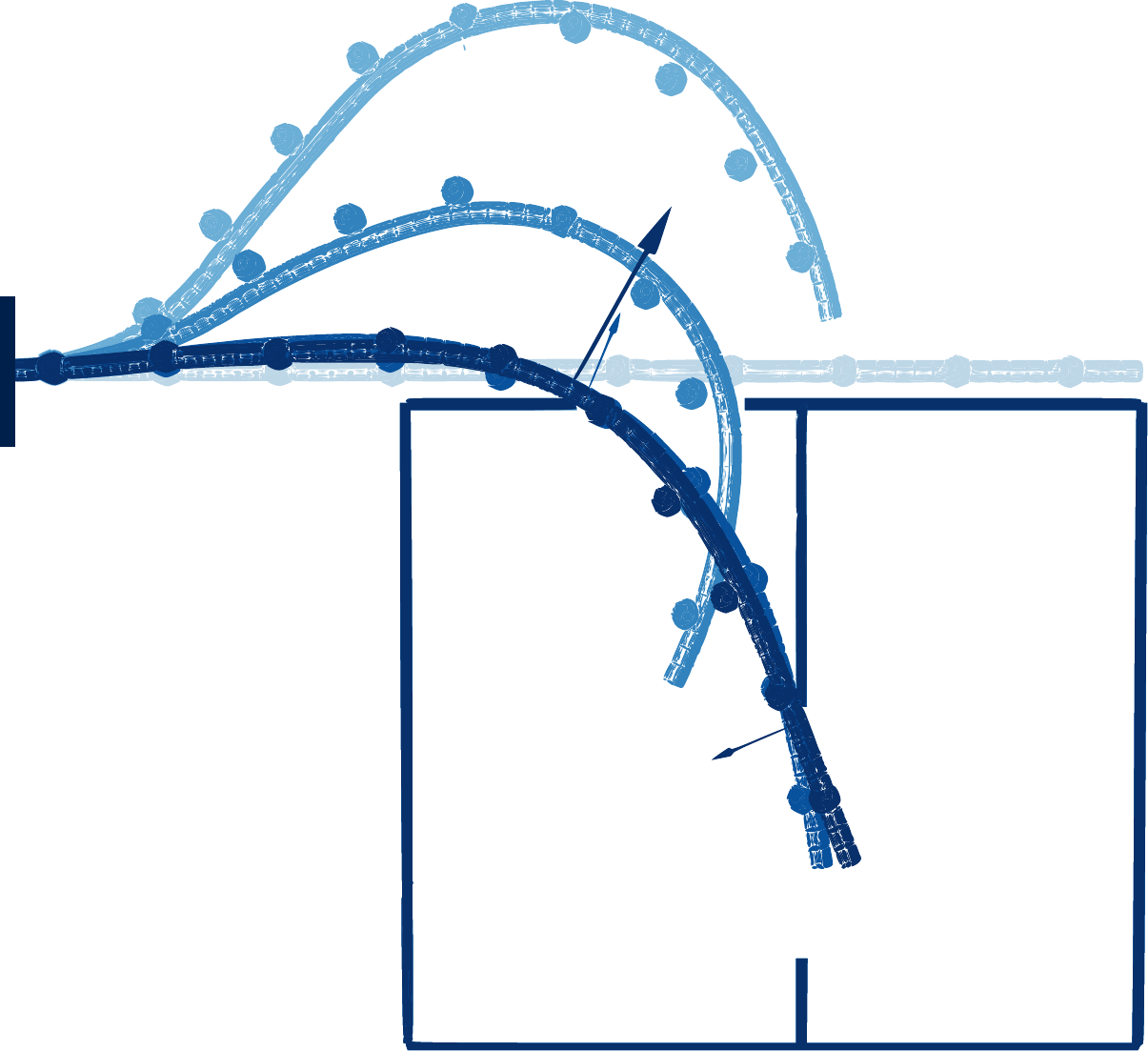}
    \caption{A simulated continuum robot navigating through a confined box environment. The estimator progressively identifies contact locations as the robot interacts with the walls.}
    \label{fig:force_estimation_box}
\end{figure}

In this section, we evaluate the proposed estimation framework during a dynamic navigation task within a confined environment. 
The simulated manipulator is driven by user-defined tendon tension inputs to navigate quasi-statically through a rigid box, as depicted in Fig.~\ref{fig:force_estimation_box}. 
For this specific experiment, we simulate a continuum robot with a total length of $L = 1 \; \mathrm{m}$, a radius of $r = 0.01 \; \mathrm{m}$, and a Young's modulus of $E = 1 \; \mathrm{MPa}$. 
The manipulator is driven by five pairs of antagonistic tendons routed along the $\pm y$-axis. The $i$-th tendon pair terminates at arc length $s_i \in \{1.0, 0.8, 0.6, 0.4, 0.2\} \; \mathrm{m}$ and is positioned at a corresponding radial offset of $d_i \in \{0.01, 0.02, 0.03, 0.04, 0.05\} \; \mathrm{m}$ from the backbone center.
While this specific navigation task is constrained to a plane for simplicity, the proposed framework is formulated in $\mathrm{SE}(3)$ and thus remains applicable to the full three-dimensional task space.

To handle an unknown number of contacts, we propose a progressive estimation algorithm leveraging the sequential nature of quasi-static collisions. 
The estimator initializes with a single Gaussian basis function. 
During operation, if the estimated force magnitude of the active basis exceeds a predefined threshold ($\tau$), the contact is registered as established, and a new basis function is dynamically appended to the state vector to capture subsequent interactions.

To ensure numerical stability and prevent degenerate solutions during optimization, we introduce two additional terms into the factor graph objective beyond those defined 
in~\eqref{eq:map_cost}.
\begin{enumerate}
    \item \textbf{Axial Force Factor.} The axial component of each basis force amplitude, in the local body frame, becomes unobservable in some configurations, particularly near-straight poses where axial loads cause minimal bending deformation. 
    Without regularization, this renders the body-frame axial component a free variable that can diverge at zero cost to the mechanics residuals. 
    We therefore apply a quadratic penalty on the tangent-direction projection of each basis amplitude:
\begin{equation}
    J_{\mathrm{axial}} = \sum_{i=1}^{M} \frac{1}{2\sigma_{\mathrm{axial}}^2}
    \bigl(\mathbf{e}_1(h(\beta_i))\t \boldsymbol{\alpha}_i\bigr)^2,
    \label{eq:axial_cost}
\end{equation}
where $\mathbf{e}_1(h(\beta_i))$ is the backbone tangent vector at $\mu_i = h(\beta_i)$ along the arc length, evaluated from the current pose estimate $\mathbf{T}_{k(\mu_i)}$. 
This factor is implemented such that its Jacobian with respect to the pose variables is strictly zero, ensuring the penalty isolates the force vector without artificially stiffening or altering the estimated backbone shape.

\item  \textbf{Spatial Separation Penalty.} When multiple basis functions are present ($M > 1$), the optimizer may collapse two or more centers $\mu_i$ to the same arc-length location, producing redundant, physically degenerate representations of a single contact point. 
To prevent this, we use a pairwise repulsion penalty on the normalized center coordinates $\mu_i = h(\beta_i)$ as follows
\begin{equation}
    J_{\mathrm{sep}} = w_\mathrm{rep}\sum_{i=1}^{M}\sum_{j=i+1}^{M}
 \exp\left(-\frac{(\mu_i - \mu_j)^2}{2\sigma_{\mathrm{rep}}^2}\right),
    \label{eq:sep_cost}
\end{equation}
 where $w_\mathrm{rep} > 0$ is a repulsion weight and $\sigma_{\mathrm{rep}}$ is the exclusion radius in normalized arc-length space. 
 This function ensures that two bases incur maximum repulsion when they approach within one basis width of each other. Note that $J_{\mathrm{sep}}$ is a non-Gaussian regularization penalty, and is included purely to enforce geometric diversity among the basis locations.
\end{enumerate}

Proper state initialization is critical for the reliable convergence of the nonlinear optimizer. 
The first time step initializes a single basis at the backbone midpoint ($s = L/2$); each subsequent step initializes from the previous step's posterior. 
When a second basis is spawned on threshold crossing, it is initialized in the distal region, halfway between the established contact and the tip ($s = 0.75L$). 
A proximal initialization instead risks a local minimum that traps the estimate until the second contact force grows large.

\begin{figure}[htbp]
    \centering
    % First Subfigure
    \begin{subfigure}[b]{0.48\textwidth}
        \centering
        \includegraphics[width=\textwidth]{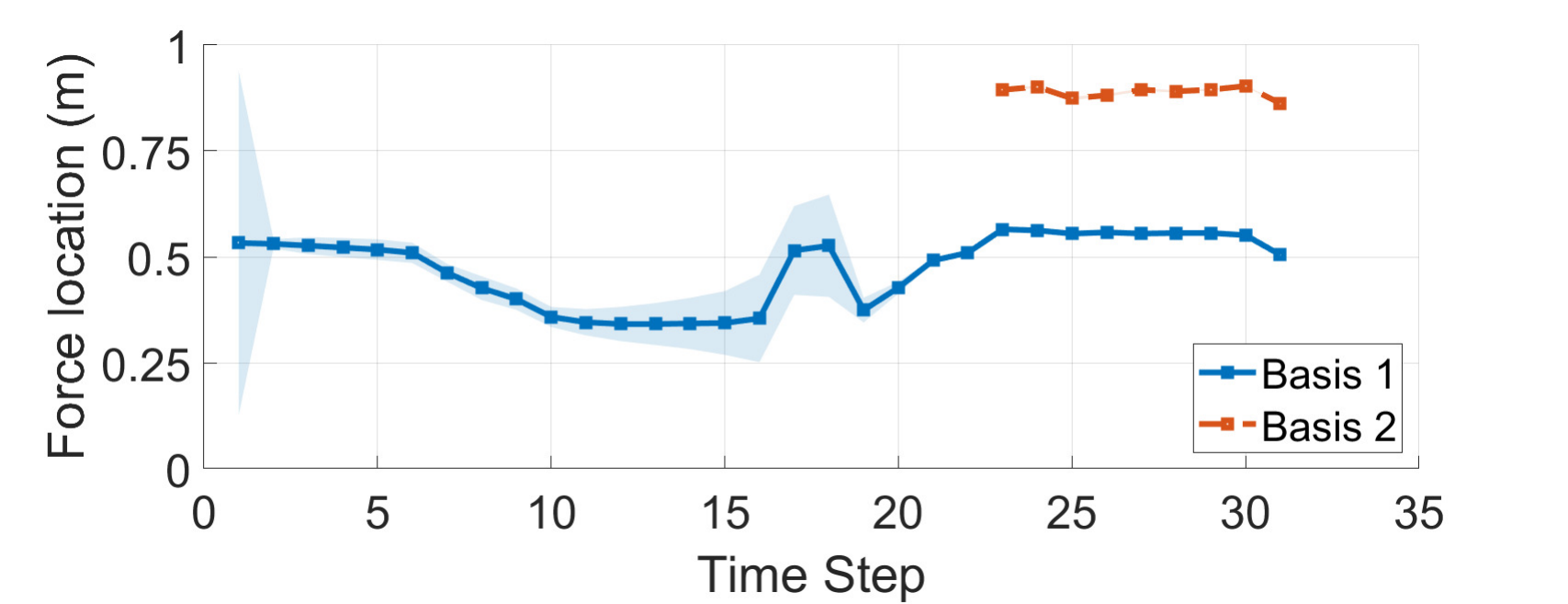}
        \caption{Estimated force locations for both basis functions.}
        \label{fig:force_loc_boxbox}
    \end{subfigure}
    \hfill % Adds flexible space between the images
    % Second Subfigure
    \begin{subfigure}[b]{0.48\textwidth}
        \centering
        \includegraphics[width=\textwidth]{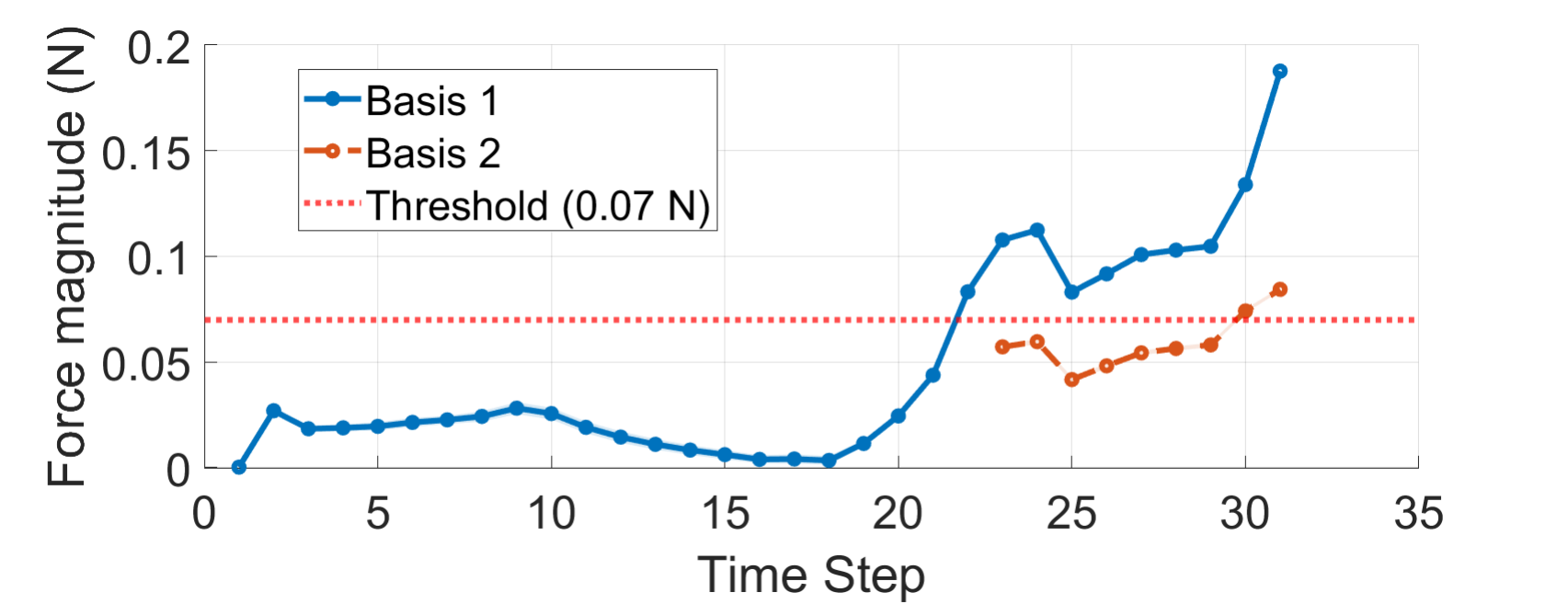}
        \caption{Estimated force magnitude for both basis functions.}
        \label{fig:force_mag_boxbox}
    \end{subfigure}
    
    \caption{Time-series analysis of the progressive multi-contact estimation scenario. A second basis function is added only after the first contact magnitude exceeds the threshold $\tau$.}
    \label{fig:boxbox}
\end{figure}

Figure~\ref{fig:boxbox} shows the estimation timeline across navigation phases. 
In free space (steps 1--20), no contact is present: the estimated magnitude stays near zero and the location fluctuates widely (Fig.~\ref{fig:boxbox}a), as the location Jacobian scales with the force amplitude and is therefore weakly observable. 
Around step 21 the robot meets the first corner; the location converges to the contact point and the magnitude rises with the applied actuation tension, crossing $\tau$ and triggering the second basis. 
The spawned basis then converges toward the tip and remains near $\tau$ as the robot engages the second wall, capturing the dual-contact state by the final steps.

\section{Conclusion}

In this paper, we have presented a unified shape and force estimation framework for tendon-driven continuum robots operating in unknown environments. 
Recognizing that discrete, node-wise force estimation is ill-posed and susceptible to ill-conditioning, we have integrated a Gaussian force parameterization into a discretized Cosserat rod mechanics model. 
By formulating the state estimation problem within a factor graph architecture, we probabilistically fused multi-modal sensor data including discrete strain, tendon tension, and pose measurements, to simultaneously estimate the robot's shape and multiple localized contact forces.

Numerical simulations showed that our method reduces mean location error compared to both a discrete node-wise baseline and an existing force-parameterization baseline, particularly under realistic sensor noise. 
Furthermore, we demonstrated a progressive estimation strategy for detecting and localizing sequential body collisions during a confined box-navigation task.

Future work will focus on integrating this estimation framework into closed-loop robotic systems, where the estimated contact locations and force magnitudes can support spatial planning, reactive obstacle avoidance, and force control. 
While the current framework relies on a quasi-static model, we also aim to extend it to full continuum dynamics for estimation during dynamic maneuvers.

\bibliographystyle{IEEEtran}

\bibliography{citations}

@article{russo2023continuum,
  title={Continuum robots: An overview},
  author={Russo, Matteo and Sadati, Seyed Mohammad Hadi and Dong, Xin and Mohammad, Abdelkhalick and Walker, Ian D and Bergeles, Christos and Xu, Kai and Axinte, Dragos A},
  journal={Advanced Intelligent Systems},
  volume={5},
  number={5},
  pages={2200367},
  year={2023}
}

@inproceedings{robinson1999continuum,
  title={Continuum robots-a state of the art},
  author={Robinson, Graham and Davies, J Bruce C},
  booktitle={IEEE International Conference on Robotics and Automation},
  pages={2849--2854},
  volume={4},
  address = {Detroit, MI, USA},
  month = {May, 10--15,},
  year={1999}
}

@article{wu2025soft,
  title={Soft growing robot explore unknown environments through obstacle interaction},
  author={Wu, Haoran and Sun, Fuchun and Huang, Canwei and Huang, Haiming and Chu, Zhongyi},
  journal={IEEE Robotics and Automation Letters},
  year={2025},
  publisher={IEEE},
  volume={10},
  number={6},
  pages={6015-6022},
}

@article{abah2021multi,
  title={A multi-modal sensor array for human--robot interaction and confined spaces exploration using continuum robots},
  author={Abah, Colette and Orekhov, Andrew L and Johnston, Garrison LH and Simaan, Nabil},
  journal={IEEE Sensors Journal},
  volume={22},
  number={4},
  pages={3585--3594},
  year={2022}
}

@article{burgner2015continuum,
  title={Continuum robots for medical applications: A survey},
  author={Burgner-Kahrs, Jessica and Rucker, D Caleb and Choset, Howie},
  journal={IEEE Transactions on Robotics},
  volume={31},
  number={6},
  pages={1261--1280},
  year={2015}
}

@article{lilge2022continuum,
  title={Continuum robot state estimation using Gaussian process regression on SE (3)},
  author={Lilge, Sven and Barfoot, Timothy D and Burgner-Kahrs, Jessica},
  journal={The International Journal of Robotics Research},
  volume={41},
  number={13-14},
  pages={1099--1120},
  year={2022}
}

@inproceedings{wang2019aero,
  title={An aero-engine inspection continuum robot with tactile sensor based on EIT for exploration and navigation in unknown environment},
  author={Wang, Yaming and Ju, Feng and Cao, Yanfei and Yun, Yahui and Wang, Yaoyao and Bai, Dongming and Chen, Bai},
  booktitle={IEEE/ASME International Conference on Advanced Intelligent Mechatronics (AIM)},
  pages={1157--1162},
  year={2019},
  address={Hong Kong, China},
  month={July 8--12,}
}

@article{yamauchi2022development,
  title={Development of a continuum robot enhanced with distributed sensors for search and rescue},
  author={Yamauchi, Yu and Ambe, Yuichi and Nagano, Hikaru and Konyo, Masashi and Bando, Yoshiaki and Ito, Eisuke and Arnold, Solvi and Yamazaki, Kimitoshi and Itoyama, Katsutoshi and Okatani, Takayuki and others},
  journal={Robomech Journal},
  volume={9},
  number={1},
  pages={8},
  year={2022}
}

@article{hu2026contact,
  title={Contact force estimation of continuum robots without embedded sensors: A review},
  author={Hu, An and Sun, Yu},
  journal={Advanced Intelligent Systems},
  volume={8},
  number={3},
  pages={e202500786},
  year={2026}
}

@article{sadati2020stiffness,
  title={Stiffness imaging with a continuum appendage: Real-time shape and tip force estimation from base load readings},
  author={Sadati, SM Hadi and Shiva, Ali and Herzig, Nicolas and Rucker, Caleb D and Hauser, Helmut and Walker, Ian D and Bergeles, Christos and Althoefer, Kaspar and Nanayakkara, Thrishantha},
  journal={IEEE Robotics and Automation Letters},
  volume={5},
  number={2},
  pages={2824--2831},
  year={2020}
}

@article{bian2025accurate,
  title={Accurate shape and tip-contact-force estimation of multisegment continuum-robotic tubular bronchoscopes},
  author={Bian, Gui-Bin and Zhang, Ming-Yang and Ye, Qiang and Ren, Han and Zhai, Yu-Peng and Ma, Ruichen and Li, Zhen},
  journal={IEEE/ASME Transactions on Mechatronics},
  volume={30},
  number={6},
  pages={6724--6735},
  year={2025}
}

@article{diezinger20223d,
  title={3D curvature-based tip load estimation for continuum robots},
  author={Diezinger, Matyas A and Tamadazte, Brahim and Laurent, Guillaume J},
  journal={IEEE Robotics and Automation Letters},
  volume={7},
  number={4},
  pages={10526--10533},
  year={2022}
}

@article{feng2021learning,
  title={A learning-based tip contact force estimation method for tendon-driven continuum manipulator},
  author={Feng, Fan and Hong, Wuzhou and Xie, Le},
  journal={Scientific Reports},
  volume={11},
  number={1},
  pages={17482},
  year={2021}
}

@article{bajo2011kinematics,
  title={Kinematics-based detection and localization of contacts along multisegment continuum robots},
  author={Bajo, Andrea and Simaan, Nabil},
  journal={IEEE Transactions on Robotics},
  volume={28},
  number={2},
  pages={291--302},
  year={2012}
}

@book{barfoot2024state,
  title={State estimation for robotics},
  author={Barfoot, Timothy D},
  year={2024},
  edition={2},
  publisher={Cambridge University Press}
}

@inproceedings{job2023multiple,
  title={Multiple-contact estimation for tendon-driven continuum robots with proprioceptive sensor information by contact particle filter and kinetostatic models},
  author={Job, Tim-David and Bensch, Martin and Schappler, Moritz},
  booktitle={IEEE/RSJ International Conference on Intelligent Robots and Systems (IROS)},
  pages={10224--10231},
  year={2023},
  address={Detroit, MI, USA},
  month={October 1--5,},
}

@article{ha2022contact,
  title={Contact localization of continuum and flexible robot using data-driven approach},
  author={Ha, Xuan Thao and Wu, Di and Lai, Chun-Feng and Ourak, Mouloud and Borghesan, Gianni and Menciassi, Arianna and Vander Poorten, Emmanuel},
  journal={IEEE Robotics and Automation Letters},
  volume={7},
  number={3},
  pages={6910--6917},
  year={2022}
}

@article{gao2024body,
  title={Body contact estimation of continuum robots with tension-profile sensing of actuation fibers},
  author={Gao, Anzhu and Lin, Zecai and Zhou, Cheng and Ai, Xiaojie and Huang, Bidan and Chen, Weidong and Yang, Guang-Zhong},
  journal={IEEE Transactions on Robotics},
  volume={40},
  pages={1492--1508},
  year={2024}
}

@article{feliu2025actuation,
  title={Actuation reading insights: Estimating shape and forces in tendon-driven slender soft robots},
  author={Feliu-Talegon, Daniel and Alkayas, Abdulaziz Y and Adamu, Yusuf Abdullahi and Mathew, Anup Teejo and Renda, Federico},
  journal={IEEE/ASME Transactions on Mechatronics},
  year={2025},
  volume={30},
  number={6},
  pages={7878-7888}
}

@article{wang2026strain,
  title={Strain-based Shape and 3D Force Estimation for Rod-driven Continuum Robots with Stretch Sensors},
  author={Wang, Peiyi and Feliu-Talegon, Daniel and Sun, Yuchen and Xie, Zhexin and Xin, Wenci and Nazeer, Muhammad Sunny and Della Santina, Cosimo and Laschi, Cecilia and Renda, Federico},
  journal={IEEE Transactions on Robotics},
   year={2026},
  volume={42},
  number={},
  pages={894-911},
}

@article{ferguson2026fast,
  author={Ferguson, James M. and Kuntz, Alan and Hermans, Tucker},
  journal={IEEE Robotics and Automation Letters}, 
  title={Continuum Robot State Estimation with Actuation Uncertainty}, 
  year={2026},
  volume={},
  number={},
  pages={1-8},
}

@article{ferguson_unified_2024,
    title = {Unified Shape and External Load State Estimation for Continuum Robots},
    volume = {40},
    journal = {IEEE Transactions on Robotics},
    author = {Ferguson, James M. and Rucker, D. Caleb and Webster, Robert J.},
    year = {2024},
    pages = {1813--1827},
}

@article{mathew2025reduced,
  title={Reduced order modeling of hybrid soft-rigid robots using global, local, and state-dependent strain parameterization},
  author={Mathew, Anup Teejo and Feliu-Talegon, Daniel and Alkayas, Abdulaziz Y and Boyer, Frederic and Renda, Federico},
  journal={The International Journal of Robotics Research},
  volume={44},
  number={1},
  pages={129--154},
  year={2025}
}

@article{dellaert2017factor,
  title={Factor graphs for robot perception},
  author={Dellaert, Frank and Kaess, Michael},
  journal={Foundations and Trends in Robotics},
  volume={6},
  number={1-2},
  pages={1--139},
  year={2017}
}

@article{aloi2022estimating,
  title={Estimating forces along continuum robots},
  author={Aloi, Vincent and Dang, Khoa T and Barth, Eric J and Rucker, Caleb},
  journal={IEEE Robotics and Automation Letters},
  volume={7},
  number={4},
  pages={8877--8884},
  year={2022}
}

@article{dellaert2012factor,
  title={Factor graphs and GTSAM: A hands-on introduction},
  author={Dellaert, Frank},
  journal={Georgia Institute of Technology, Tech. Rep},
  volume={2},
  number={4},
  year={2012}
}

@incollection{powell1970hybrid,
  title={A hybrid method for nonlinear equations},
  author={Powell, Michael J. D.},
  booktitle={Numerical Methods for Nonlinear Algebraic Equations},
  editor={Rabinowitz, Philip},
  pages={87--114},
  year={1970},
  publisher={Gordon and Breach Science Publishers},
  address={London}
}

\appendix[Analytical Jacobian of the Wrench Residual with Respect to the Force Vector Parameters]
\label{app:jacobian}

\numberwithin{equation}{section}
\setcounter{equation}{0}

The wrench-balance residual \eqref{eq:wrench_residual} depends on the force parameters $\boldsymbol{\gamma}$ solely through the cumulative interval force $\Delta\mathbf{v}_{e,k} \triangleq \mathbf{v}_{e,k+1} - \mathbf{v}_{e,k}$. Therefore, we have
\begin{equation}
    \frac{\partial \mathbf{e}_{\Lambda,k}}{\partial \boldsymbol{\gamma}}
    = \underbrace{\frac{\partial \mathbf{e}_{\Lambda,k}}{\partial \Delta\mathbf{v}_{e,k}}}_{6\times 3}
      \cdot
      \underbrace{\frac{\partial \Delta\mathbf{v}_{e,k}}{\partial \boldsymbol{\gamma}}}_{3\times 4M},
\end{equation}
where
\begin{equation}
    \frac{\partial \mathbf{e}_{\Lambda,k}}{\partial \Delta\mathbf{v}_{e,k}} =  \begin{bmatrix}\mathbf{0}_{3\times3}\\ \mathbf{R}_{k+1}\t\end{bmatrix}.
\end{equation}
From \eqref{eq:v_def}, the integrated interval force is block-separable across the $M$ basis functions,
\begin{equation}
    \Delta\mathbf{v}_{e,k}(\gamma)
        = \sum_{i=1}^{M} \boldsymbol{\alpha}_i\, \Phi_{i,k},
\end{equation}
where
\begin{equation}
    \Phi_{i,k} = \frac{1}{2}\!\left[
        \mathrm{erf}\!\left(\frac{s_{k+1}-\mu_i}{\sqrt{2}\,\sigma_i}\right)
        - \mathrm{erf}\!\left(\frac{s_{k}-\mu_i}{\sqrt{2}\,\sigma_i}\right)
    \right]
\end{equation}
with $\mu_i = \tfrac{L}{2}\bigl(1+\tanh\beta_i\bigr)$ from \eqref{eq:mu_param}. 
The amplitude derivative is immediate from $\partial \Delta\mathbf{v}_{e,k}/\partial \boldsymbol{\alpha}_i = \Phi_{i,k}\, I_3$. 
Using the chain rule, we get 
\begin{equation}
        \frac{\partial \Delta\mathbf{v}_{e,k}}{\partial\beta_i} =
\boldsymbol{\alpha}_i\,\frac{\partial \Phi_{i,k}}{\partial \mu_i} \, \frac{\partial \mu_i}{\partial \beta_i},
\end{equation}
where
\begin{equation}
    \frac{\partial \Phi_{i,k}}{\partial \mu_i}
        = \frac{e^{-z_{i,k}^2} - e^{-z_{i,k+1}^2}}{\sqrt{2\pi}\,\sigma_i},
\end{equation}
and
\begin{equation}
    \frac{\partial \mu_i}{\partial \beta_i}
        = \frac{L}{2}\,\mathrm{sech}^2(\beta_i),
\end{equation}
with $z_{i,k} = (s_k-\mu_i)/(\sqrt{2}\,\sigma_i)$.

Because the derivative with respect to $\beta_i$ is proportional to $\boldsymbol{\alpha}_i$, a zero-force amplitude renders the location parameter unobservable, zeroing out its Jacobian block. 
To prevent this, we initialize each amplitude with a small magnitude $\epsilon > 0$. Stacking the partial derivatives for all $M$ basis functions yields
\begin{equation}
    \frac{\partial \Delta\mathbf{v}_{e,k}}{\partial \boldsymbol{\gamma}}
    = \Bigl[\ \cdots\ \big|\Phi_{i,k} I_3|\
        \boldsymbol{\alpha}_i\,\frac{\partial\Phi_{i,k}}{\partial\beta_i} \ \big|\ \cdots\ \Bigr],
\end{equation}
so that
\begin{equation}
    \frac{\partial \mathbf{e}_{\Lambda,k}}{\partial \boldsymbol{\gamma}}
    = \begin{bmatrix}
        \mathbf{0}_{3\times 4M}\\[2pt]
        \mathbf{R}_{k+1}\t\,\dfrac{\partial \Delta\mathbf{v}_{e,k}}{\partial \gamma}
      \end{bmatrix} \in \mathbb{R}^{6\times 4M}.
\end{equation}

\end{document}